%% file: iclr2023_conference_tinypaper.tex
\documentclass{article} 
\usepackage{iclr2023_conference_tinypaper,times}
\usepackage{booktabs}
\usepackage{graphicx}
\usepackage{subcaption}
\input{math_commands.tex}

\usepackage{hyperref}
\usepackage{url}

\title{Parameter \& Data Efficient Spectral Style-DCGAN}

\author{Aryan Garg \\
\texttt{aryangarg019@gmail.com} \\
}

\iclrfinalcopy 

\begin{document}

\maketitle

\begin{abstract}
We present a simple, highly parameter, and data-efficient adversarial network for unconditional face generation. Our method: Spectral Style-DCGAN or SSD utilizes only 6.574 million parameters and 4739 dog faces from the Animal Faces HQ (AFHQ)~\citep{choi2020starganv2} dataset as training samples while preserving fidelity at low resolutions up to 64x64. Code available at  \href{https://github.com/Aryan-Garg/StyleDCGAN}{Github}. 
\end{abstract}

\begin{figure}[ht]
     \centering
     \includegraphics[width=0.99\linewidth]{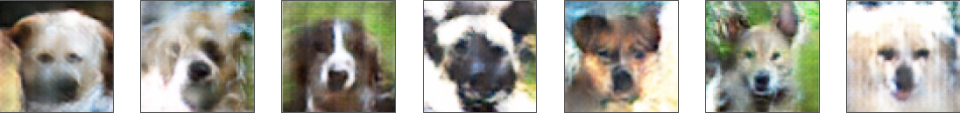}
    \caption{Spectral normalized discriminator improvement synthesis}
    \label{fig:main_teaser}
\end{figure}

\section{Introduction}
\label{sec:intro}
Typically, modern high-fidelity generators need $\sim$ $10^5 - 10^6$ images, which makes training GANs extremely costly and disadvantages important sectors like medicine~\citep{medicine}.  
DCGAN~\citep{dcgan} the first and arguably the smallest convolutional GAN~\citep{goodfellow2014generative} still required tremendous amounts of training data: LSUN~\citep{yu15lsun} ($\sim$3 million) and ImageNet-1K~\citep{ILSVRC15} ($\sim$1.28 million). 
Data-efficient GAN frameworks that utilize tiny datasets of sizes $\sim10^3$ are often marred with extensive compute requirements and/or massive parameterization~\citep{Tran_2021, lu2023machine, styleGAN2}. See \ref{sec:intro2}. To alleviate the high data and parameters, we make the following contributions: a) We contribute a robust GAN framework that utilizes tiny datasets ($\sim$5000 samples), comparable to~\citep{styleGAN2, kumari2021ensembling} while being extremely parameter-efficient. At 6.57 million parameters we use 624.44\% fewer parameters than the state-of-the-art StyleGAN and variants. b) We demonstrate for the first time how spectral normalization implicitly aids meaningful generator-side learning and latent space disentanglement. 


\section{Method}
\label{method}
To effectively understand the distribution and consequently not overfit, the generator needs to learn diverse image contents(pose, identity) and styles(hair, eye color). 
Gatys~\citep{gatys_content_and_style_separation} show that style and content are separable. 
Furthermore, sufficient disentanglement in the stochastic style space is needed for the generator to localize the style to the relevant content or regions of the image. 
Inspired by StyleGAN's~\citep{styleGAN} learned style constant and adaptive instance normalization (adaIN)~\citep{adaIN} to enforce the style, our generator's head is a tiny 100-dimensional 4-layer MLP head, unlike StyleGAN-2's huge 512-dimensional 8-layer MLP. 
This head maps the noise to a disentangled style-space similar to~\cite{styleGAN, styleGAN2}. 
The learned style vector is used for adaIN layers~\citep{huang2017arbitrary} introduced in our custom-DCGAN's deconvolutional upsampling generator block.
Finally, the synthesized and real images are sent to a spectrally normalized (SN)~\cite{spectral_why} weights' ($W_{SN}:= W / \sigma(W)$) custom-DCGAN discriminator. 
This discriminator choice is inspired by the key insights from~\cite{key_1, key_2, style_ADA} that the discriminator overfits in low data regimes to the training examples subsequently making the feedback to the generator meaningless. 
This diverges the overall training. 
To avoid vanishing gradients and increasing the parameters significantly we choose spectral normalization as an internal regularizer for the discriminator. 
Overall, these potent improvements allow our method to be data and parameter-efficient.
Refer to Appendix~\ref{sec:arch_trainDeets} for more architectural details.

     
         

\section{Experiments}
\subsection{Results \& Spectral Normalization Ablation}
\begin{figure}[ht]
     \centering
     \includegraphics[width=0.99\linewidth]{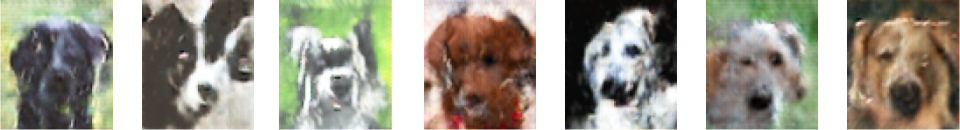}
    \caption{Ablation: No spectral normalization. Style Mapping and AdaIN synthesis}
    \label{fig:map_adaIN}
\end{figure}

We observe that SN discriminator improvement slows the discriminator (Fig.~\ref{fig:ablation_sn}) down thus allowing the generator to learn meaningful and coherent faces.
 See Fig~\ref{fig:main_teaser}. Without the SN improvement, the generator is not able to learn better high-level facial attributes like head shape, number of ears, etc. as shown in the ablation Fig~\ref{fig:map_adaIN}. 

\subsection{Quantitative Comparison \& Data-Efficiency Ablation}
We quantitatively evaluate our method, DCGAN~\citep{dcgan} and StyleGAN~\citep{styleGAN} under identical deterministic settings using FID~\citep{heusel2018gans_fid} in Table~\ref{tab:main_comparison}. We trained all baselines and our method for 20 epochs on the AFHQ Dog Subset with default hyperparameters.
\begin{table}[h]
    \begin{center}
    \begin{tabular}{lcc}
    \toprule
         Method & Parameters (M) & FID $\downarrow$ \\
    \midrule
         DCGAN~\citep{dcgan} & \textbf{6.342} & 424.512 \\  
         StyleGAN~\citep{styleGAN} & 47.625 & 355.620 \\  
    \midrule
         Ours & 6.574 & \textbf{285.579} \\
    \bottomrule
    \end{tabular}
    \caption{Generation fidelity and parameters: Our method produces the highest-fidelity results and uses 624.44\% fewer parameters compared to StyleGAN.  We also observe that StyleGAN requires significantly more data ($\sim 5 \times 10^4$ images) and intensive training time to match our fidelity score.}
    \label{tab:main_comparison}
    \end{center}
\end{table}
To further validate our data-efficient solution, we perform ablation in Table~\ref{tab:data_size} and evaluate the generation fidelity. Due to the aliased resizing operation, we observe an inconsistency in PyTorch-FID scores, similar to~\citep{parmar2021cleanfid}. So, we evaluate further using clean-FID~\citep{parmar2021cleanfid} and clean-Kernel Inception Distance~\citep{kid, parmar2021cleanfid}
\begin{table}[h]
    \begin{center}
    \begin{tabular}{lccc}
    \toprule
         Training Data ($K$\%) & FID $\downarrow$ & clean-FID $\downarrow$ & clean-KID $\downarrow$ \\
    \midrule
        100\%  & 285.579 & \textbf{274.139} & \textbf{0.2054} \\  
         75\% & \textbf{241.581} & 283.536 & 0.2379 \\  
         50\% & 294.467 & 293.038  & 0.2452 \\ 
         25\% & 319.899 & 319.241 & 0.2510 \\    
    \bottomrule
    \end{tabular}
    \caption{Ablation: Data Efficiency. Using $K$\% of the dogs-AFHQ training split, we estimate the generation quality using FID, clean-FID~\citep{parmar2021cleanfid} and clean Kernel Inception Distance~\citep{kid} (clean-KID)}
    \label{tab:data_size}
    \end{center}
\end{table}

\newpage

\section*{URM Statement}
The authors acknowledge that the author of this work meets the URM criteria of the ICLR 2024 Tiny Papers Track.

\bibliography{iclr2023_conference_tinypaper}
\bibliographystyle{iclr2023_conference_tinypaper}

\appendix
\section{Appendix}
\label{appendix}
The code is available at: \href{https://github.com/ANonyMouxe/StyleDCGAN}{repo}.

\subsection{Important note on the FID scores} 
FID scores are high as we upsampled the generated images while downsampling the real distribution from 512x512 due to compute limitations. Additionally, the 2048th layer of InceptionV3 is used to compute FID. 

We notice the counter-intuitive dip in FID when using only 75\% data in Tab.~\ref{tab:data_size} and pin-point the issue to large aliasing during resizing operations done internally by PyTorch-FID, as first observed in~\citep{parmar2021cleanfid}. 
So, we decide to use the clean-FID and clean-KID metrics from~\citep{parmar2021cleanfid} as well for Tab.~\ref{tab:data_size}. 

The StyleDCGAN notebook in our \href{https://github.com/ANonyMouxe/StyleDCGAN}{codebase} has the output cells saved as well with the clean-FID metrics for verification.

\subsection{Extended Introduction/Motivation}
\label{sec:intro2}
With the substantial fidelity gains of GANs~\cite{goodfellow2014generative} over other explicit density modeling methods including VAEs~\citep{vae}, many ground-breaking applications in defense~\citep{military}, engineering~\citep{civil_eng}, healthcare~\citep{medicine} have made GANs mainstream. 

GANs are being succeeded by diffusion and Poisson generative models~\citep{diffusion} but still are extremely valuable as they use significantly less compute while still staying competitive in terms of fidelity~\citep{dhariwal2021diffusion}. 

However, modern GANs still require massive (of the order of $10^6$) training datasets~\citep{style_ADA}, and are often parametrically overburdened. For instance, StyleGAN-XL~\citep{sauer2022styleganxl} used $3\times$ more parameters than StyleGAN3~\citep{style3} and consumed massive amounts of energy during training. The environmental impact of StyleGAN3 was 224.62 Mega Watt Hours(MWh) and 91.77 GPU years or the time it would take one NVIDIA V100 to train. To put the energy impact in perspective, an average American citizen uses a total of 12.71 MWh in a year (data from EIA report, 2022).

This significantly limits the usage of GANs to a select few. When small GANs (networks' parameters) and low-data regimes($\sim10^3$) are used, detrimental issues like overfitting, mode collapse, aggravated loss divergence due to the inherent instability in the adversarial training set-up and significantly reduced generation fidelity are observed~\citep{de_GAN}. This is a critical issue for deployment and crucial applications. Especially in the healthcare sector, which is plagued by a chronic scarcity of labeled data~\citep{medicine}.

To alleviate these pressing issues and still retain the advantages of this valuable technology, we present a parameter and data-efficient GAN that also trains quickly and stably.

\subsection{Related Works}
\label{sec:related_works}
\begin{table}[h]
    \centering
    \begin{tabular}{l|c}
    \toprule
        \textbf{Strategy} & \textbf{Methods} \\    
    \midrule
         Data Augmentation &  Differential Augmentation~\citep{zhao2020differentiable}, \\   
         & GANs for synthetic data~\citep{math_tree}\\  
         Loss side & WGAN~\citep{arjovsky2017wasserstein}, Vision-Aided GAN~\cite{kumari2021ensembling}\\  
         & WGAN-GP, RaGAN~\citep{ragan} \\   
         Training Regimes & TTUR~\citep{ttur} \\
         Network Side & StyleGAN2-ADA~\citep{style_ADA}, \textbf{Ours} \\     
    \bottomrule
    \end{tabular}
    \caption{Related Works: In this table, we present noteworthy works for increasing data-efficiency(DE), parameter-efficiency(PE), and overall training stability(TS) of GANs}
    \label{tab:rW}
\end{table}

\subsection{Dataset}
\label{sec:dataset}
We choose the dogs' subset of the AFQH~\citep{choi2020starganv2} dataset and use the default StarGAN-v2 split where 4739 are training samples while 500 are reserved for validation. We use PyTorch's default bilinear interpolation to downsample all images to 64x64 from the original 512x512 resolution, for training and validation.

\begin{figure}[h]
    \centering
    \includegraphics[width=0.9\textwidth]{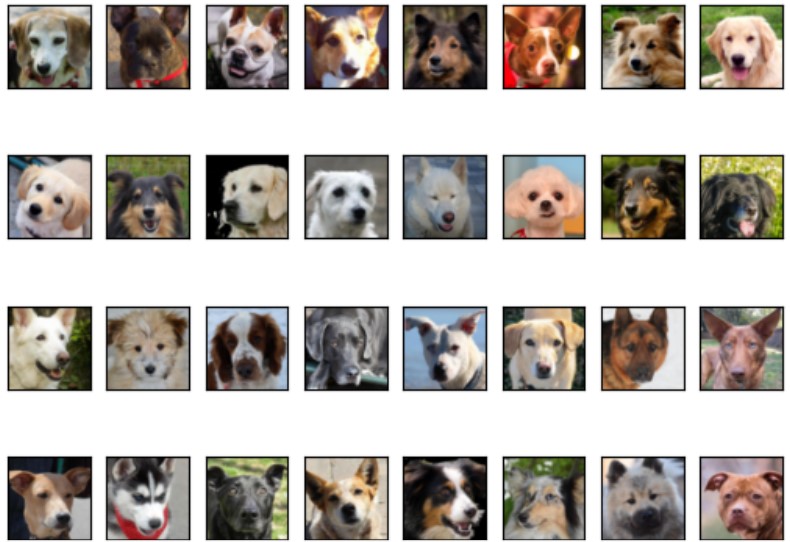}
    \caption{Training dataset samples at original 512x512 resolution.}
    \label{fig:samples_training_set}
\end{figure}

\subsection{Architecture \& Training Details}
\label{sec:arch_trainDeets}
\subsubsection{Generator}
The Generator takes in a 100-dimensional noise vector ($Z$-space) and then is mapped to another 100-dimensional vector ($W$-space) using a custom MLP named: MappingNetwork. This transformed vector aids in the disentanglement of the feature space and acts as a style vector used for computations for the adaptive instance normalization layers in the generator. Nevertheless, the MappingNetwork consists of 4 linear layers (same input-output dimensionality), each followed by an in-place leaky-ReLU activation with a slope of 0.2. 

Following this simple mapping, we move towards a custom DCGAN where each fractionally strided convolution or transposed convolution is followed by an adaptive instance normalization~\citep{adaIN} (just like StyleGAN) and finally a ReLU activation. The adaptive instance normalization layer repeatedly uses the W-space transformed latent to learn an affine transformation for the already instance normed output of the transposed-convolution layer. Simply put, $AdaIN(x_i, w)  = \alpha(w) * \frac{(x_i - \mu_i)}{\sqrt{\sigma^2_i + \epsilon}} + \beta(w)$ where $\alpha$ and $\beta$ denote learnable linear layers.

The last and the 5th transposed convolution layer is special as it does not have an adaptive instance normalization layer and is succeeded by a TanH activation layer. These changes mark a significant deviation in the generator from both StyleGAN~\citep{styleGAN} and DCGAN~\citep{dcgan}.

The generator contains a total of 3,809,104 parameters and is trained with an Adam~\citep{kingma2017adam} optimizer with a learning rate of $0.0005$ and $\beta_1$ of $0.5$ and $\beta_2$ equal to $0.999$.

\subsubsection{Discriminator}
Our discriminator is a tiny (2,765,568 parameters) 5-convolution layer CNN that uses spectral normalization followed by standard batch normalization and in-place Leaky-ReLU with default PyTorch parameters. This orthogonal combination of spectral normalization and then batch normalization layers provides enough regularization for stable adversarial training with the Adam~\citep{kingma2017adam} optimizer. The learning rate is $0.0005$ and $\beta_1$ is set to $0.5$ while $\beta_2$ equals $0.999$. 

\subsubsection{Alternate Steps, Criterion \& Framework}
For every generator step, we move the discriminator one step as well. We only use the binary cross-entropy or adversarial criterion ($L_adv(y_i, x_i)$ = $y_i * log(x_i) + (1 - y_i) * log(1 - x_i)$; $y_i \in \{0, 1\}$) for our unconditional generative solution. We monitor the generator's validation loss and save the best one. We use PyTorch-Lightning for abstraction and a free Google Colab P100-GPU session for 20 epochs of training with a batch size of 32.

Please refer to the notebook version of our implementation in \href{https://github.com/ANonyMouxe/StyleDCGAN}{repo} for a detailed architecture summary, training parameters, and more details.

\subsection{Further Experiments}
\subsubsection{Ablation: With and without spectral normalization training losses}
\begin{figure}[h]
    \centering
    \includegraphics[width=0.9\textwidth]{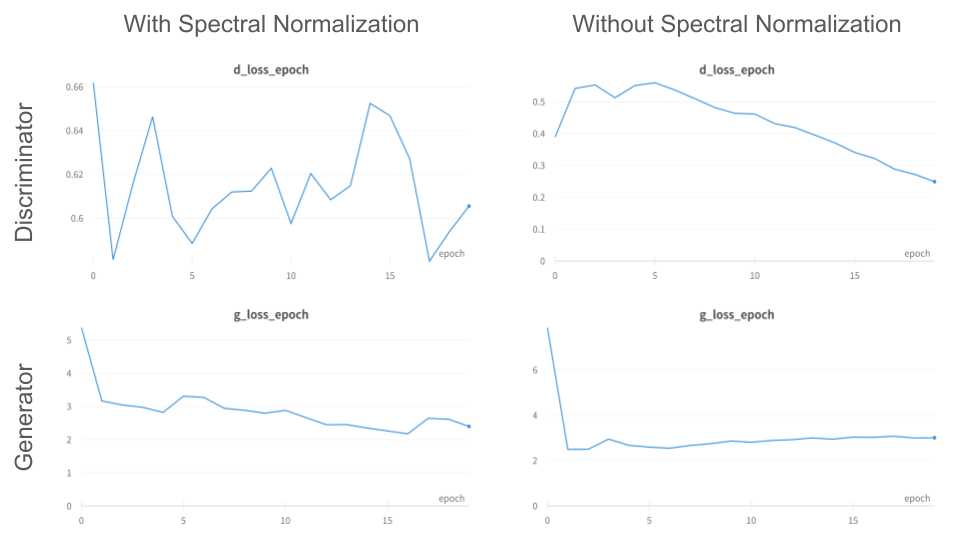}
    \caption{Ablation study: Effect of spectral normalization layers in the discriminator. The left column represents training loss plots of the discriminator(top) and generator(bottom) with the spectral normalization layers in the discriminator; while the right column demonstrates the ablated-discriminator-training loss plots.}
    \label{fig:ablation_sn}
\end{figure}

We argue that the weight regularization or spectral normalization slows down the discriminator allowing the generator to learn a richer and more diverse distribution. This slow learning for the generator is essential to avoid mode-collapse and diversity issues. Our argument is validated by the ablation in Fig.~\ref{fig:ablation_sn}

\subsubsection{Ablation: Style-Mapping Network Only Results (No AdaIN)}
Here we see that disentanglement is not enough. We additionally need to repetitively fuse the learned style vector to get consistent textures, hair, and facial features and remove the aberrations. This is essential to fix the known problem that the latent space is always continuous, however, the real dataset is always discrete and discontinuous.

\begin{figure}[ht]
     \centering
     \includegraphics[width=0.99\linewidth]{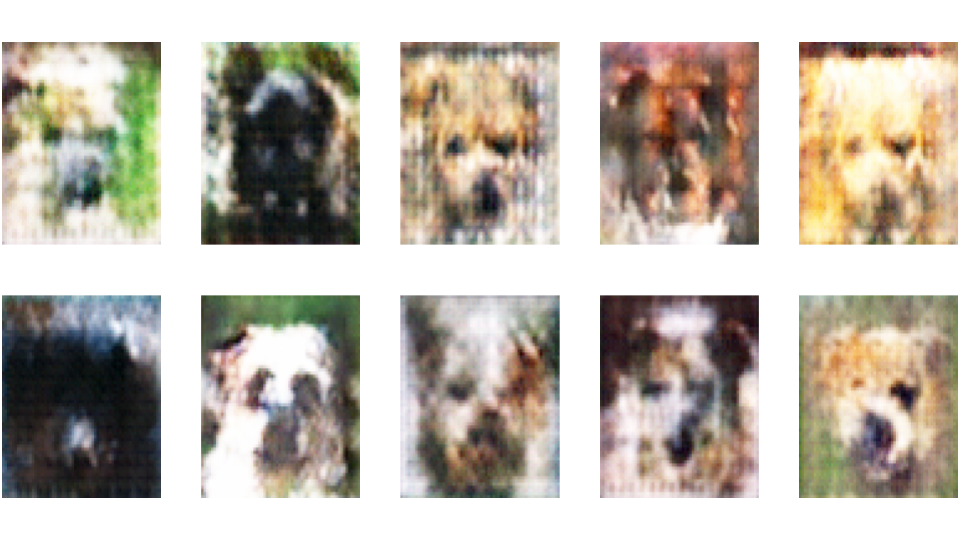}
    \caption{Ablation: Disentangling the latent vectors. Style-Mapping without AdaIN Synthesis}
    \label{fig:mapper_only}
\end{figure}

\subsection{More Qualitative Results for Ablation: Without Spectral Normalization}
We demonstrate some additional results of the ablation presented in Fig.\ref{fig:map_adaIN}.
\begin{figure}[ht]
     \centering
     \includegraphics[width=0.99\linewidth]{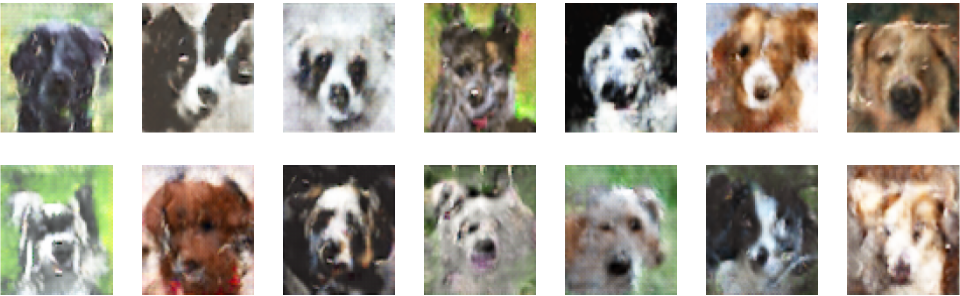}
    \caption{More Results of the ablation: No spectral normalization. Style Mapping and AdaIN synthesis}
    \label{fig:map_adaIN_2}
\end{figure}

\subsection{Z-Space Interpolation}
\begin{figure}[ht]
     \centering
    \begin{subfigure}[b]{0.1\textwidth}
         \centering
         \includegraphics[scale=0.9]{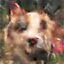}
         \caption{$z_1$}
    \end{subfigure}%
    \hfill
    \begin{subfigure}[b]{0.1\textwidth}
         \centering
         \includegraphics[scale=0.9]{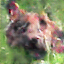}
         \caption{}
    \end{subfigure}%
    \hfill
    \begin{subfigure}[b]{0.1\textwidth}
         \centering
         \includegraphics[scale=0.9]{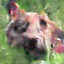}
         \caption{}
    \end{subfigure}%
    \hfill
    \begin{subfigure}[b]{0.1\textwidth}
         \centering
         \includegraphics[scale=0.9]{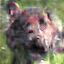}
         \caption{$z_2$}
    \end{subfigure}%
    \caption{Z-space Interpolation. We manually find a direction along which the ears are preserved. All dogs have just the left ear standing tall in (a) - (d).} 
    \label{fig:zspace_interp}
\end{figure}

We find a feature-preserving direction in our generator and linearly interpolate between two vectors sampled from this space. 
See~\ref{fig:zspace_interp}.

\subsection{Conclusion}
We present a highly parameter and data-efficient GAN framework inspired by the seminal works: DCGAN and StyleGAN; getting the best of both worlds and further regularizing it with spectral normalization to learn meaningful and coherent underlying data distributions.
Our work faces the limitation of not being able to impose any facial priors like lateral symmetry, pose, and feature positioning on the unconditional generation. Attention layers, as proposed in SAGAN~\citep{sagan}, can alleviate the same and aid in higher color consistency across the output image. However, this comes at an additional computational cost.  

\end{document}

%% file: math_commands.tex
\usepackage{amsmath,amsfonts,bm}

\def\eqref#1{equation~\ref{#1}}

\def\1{\bm{1}}

\DeclareMathAlphabet{\mathsfit}{\encodingdefault}{\sfdefault}{m}{sl}
\SetMathAlphabet{\mathsfit}{bold}{\encodingdefault}{\sfdefault}{bx}{n}

